\pdfoutput=1

\documentclass[11pt]{article}

\usepackage{EMNLP2023}

\usepackage{times}
\usepackage{latexsym}

\usepackage[T1]{fontenc}

\usepackage[utf8]{inputenc}

\usepackage{microtype}

\usepackage{inconsolata}
\usepackage{graphicx}
\usepackage{amsmath}
\usepackage{booktabs}

\usepackage{bm}
\usepackage{makecell}
\usepackage{float}
\usepackage{subcaption}
\usepackage{footmisc}

\title{\textsc{ProMiNet}: Prototype-based Multi-View Network for Interpretable Email Response Prediction}

\author{Yuqing Wang$^*$ \\
  University of California, Santa Barbara \\
  Santa Barbara, CA 93106 \\
  \texttt{wang603@ucsb.edu} \And
  Prashanth Vijayaraghavan$^*$\\
  IBM Research\\
  San Jose, CA 95120\\
  \texttt{prashanthv@ibm.com} \And
  Ehsan Degan\\
  IBM Research\\
    San Jose, CA 95120\\
  \texttt{edehgha@us.ibm.com}
  }

\begin{document}
\maketitle
\def\thefootnote{*}\footnotetext{These authors contributed equally to this work.}\def\thefootnote{\arabic{footnote}}
\begin{abstract}

Email is a widely used tool for business communication, and email marketing has emerged as a cost-effective strategy for enterprises. While previous studies have examined factors affecting email marketing performance, limited research has focused on understanding email response behavior by considering email content and metadata. This study proposes a \textbf{Pro}totype-based \textbf{M}ulti-v\textbf{i}ew \textbf{Net}work (\textsc{ProMiNet}) that incorporates semantic and structural information from email data. By utilizing prototype learning, the \textsc{ProMiNet} model generates latent exemplars, enabling interpretable email response prediction. The model maps learned semantic and structural exemplars to observed samples in the training data at different levels of granularity, such as document, sentence, or phrase. The approach is evaluated on two real-world email datasets: the Enron corpus and an in-house Email Marketing corpus. Experimental results demonstrate that the \textsc{ProMiNet} model outperforms baseline models, achieving a $\sim3\%$ improvement in $F_1$ score on both datasets. Additionally, the model provides interpretability through prototypes at different granularity levels while maintaining comparable performance to non-interpretable models. The learned prototypes also show potential for generating suggestions to enhance email text editing and improve the likelihood of effective email responses. This research contributes to enhancing sender-receiver communication and customer engagement in email interactions.
\end{abstract}

\section{Introduction}
\begin{figure}[htbp]
\centering
\includegraphics[width=0.35\textwidth]{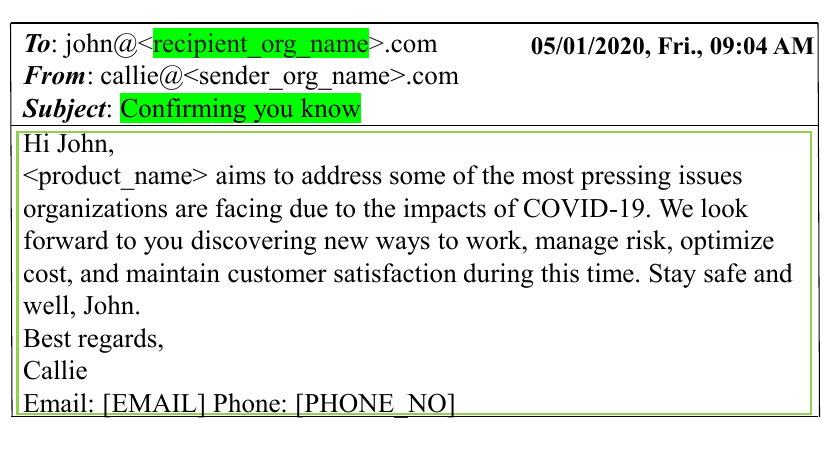}      
\caption{Sample Email with relevant contents.}
\label{fig:example_email}
\end{figure}

With the ever-increasing volume of emails being exchanged daily, email communication remains a cornerstone of business interactions and an effective means of content distribution. As the primary communication tool for organizations and individuals alike, email marketing has maintained its popularity over the years, evolving and expanding alongside advancements in technology. This form of marketing enables businesses to tailor targeted messages to customers based on their preferences, leveraging the quick, easy, and cost-effective nature of email communication. In this context, predicting customer response behavior in email marketing campaigns becomes crucial for optimizing customer-product engagements and enhancing communication efficiency between senders and recipients. Consider the example email shown in Figure~\ref{fig:example_email}, where various factors such as the email's contents (subject and body) and the recipient's organization, can influence the likelihood of receiving a response. Therefore, understanding the impact of these factors and their correlation with email response behavior is paramount. Research \cite{kim2016compete} has shown that a single word can make a substantial difference in how a text is interpreted. This insight applies to our email response prediction task, making it essential to address this challenge. The likelihood of an email receiving a response can be influenced by various factors, including the use of power words or phrases, the persuasiveness of the text, and alignment with client preferences. Given the sensitivity of words or phrases in our task, we need methods to extract both the structural and semantic information from email text to develop an effective prediction model.
Recently, there have been efforts to study different explanation techniques for text classification. These methods typically fall into two categories: post-hoc explanation methods ~\cite{madsen2021post} and self-explaining approaches ~\cite{alvarez2018towards}. Post-hoc explanations use an additional explanatory model to provide explanations after making predictions, while self-explaining approaches generate explanations simultaneously with the prediction. However, post-hoc explanations may not accurately reveal the reasoning process of the original model ~\cite{rudin2019stop}, making it preferable to build models with inherent interpretability. In this work, we propose \textsc{ProMiNet}, a novel interpretable email response prediction model that integrates semantic and structural information from email data. \textsc{ProMiNet} utilizes prototype learning, a form of case-based reasoning, to make predictions based on similarities to representative examples (prototypes) in the training data. Unlike existing prototype-based architectures, \textsc{ProMiNet} provides explanations from multiple perspectives: semantic (using transformer-based models) and structural (using graph-based dependency parsing with GNN). By leveraging a multi-branch network, \textsc{ProMiNet} offers holistic explanations at different levels, including document-level, sentence-level, and phrase-level prototypes. We conduct quantitative analyses and ablation studies using two real-world email datasets: the Enron corpus and the in-house email marketing corpus. Our \textsc{ProMiNet} model achieves superior performance and offers explanations that simulate potential edits, resulting in improved response rates.
\textbf{Contributions:} The key contributions of this work are summarized as follows: 
\begin{itemize}
    \item We present \textsc{ProMiNet}, the inaugural method for interpretable email response prediction. By combining transformer-based models and dependency graphs with GNN, our approach captures semantic and structural information at various granularities.
    \item We conduct extensive experiments on real-world email corpora. \textsc{ProMiNet} outperforms the strongest baselines on both the Enron and Email Marketing corpus.
    \item Simulation experiments demonstrate the effectiveness of learned prototypes in generating email text editing suggestions, leading to a significant enhancement in the overall email response likelihood. These results indicate promising avenues for further research.
\end{itemize}

\section{Related Work}\label{related_work}

\subsection{Email Response Prediction}
Researchers have used machine learning methods to improve email efficiency by predicting email responses. Previous work includes predicting email importance and ranking by likelihood of user action~\cite{aberdeen2010learning}, classifying emails into common actions -- read, reply, delete, and delete-WithoutRead ~\cite{di2016you}, and characterizing response behavior based on various factors~\cite{on2010mining,kooti2015evolution,qadir2016activity} including time, length, and conversion, temporal, textual properties, and historical interactions. Our work differs from previous studies by considering both semantic and structural information in email response prediction and developing an interpretable model.

\subsection{Explainability in Text Classification}
Model explainability has gained significant attention with different explainability methods categorized into post-hoc or self-explaining. Post-hoc methods ~\cite{ribeiro2016should,simonyan2013deep,smilkov2017smoothgrad, arras2016explaining} separate explanations from predictions, while self-explaining methods \cite{bahdanau2014neural,rajagopal2021selfexplain} generate explanations simultaneously with predictions. Drawing from previous studies \cite{sun2020self,ming2019interpretable}, our work falls into the self-explainable category, providing explanations through prototypes.
Prototype-based networks make decisions based on the similarity between inputs and selected prototypes. Originally used for image classification~\cite{chen2019looks}, several methods ~\cite{ming2019interpretable,hong2020interpretable,plucinski2021prototypical} have been adapted for text classification, where a similarity score is used to learn prototypes, that represent the characteristic patterns in the data. These prototypes serve as exemplars or representative instances from the dataset. However, these models providing unilateral explanations have limitations as they lack granularity, provide an incomplete picture, have limited coverage, and reduced interpretability. In contrast, granular prototypes produced by our \textsc{ProMiNet} offer a more nuanced and interpretable approach to understanding email data.

\section{Problem Setup}
We tackle the interpretable email response prediction problem as a self-explainable binary classification task. Given a training set $\mathcal{D}$ with email texts $x_i$ and binary response labels $y_i \in \{0,1\}$, our goal is to predict the likelihood of receiving a response while providing insights into the decision process. The labels indicate whether an email received a response (1) or not (0), which could include clicks, views, or replies. To enhance interpretability, we learn latent prototypes at the document, sentence, and phrase levels, mapping them to representative observations in the training set. These prototypes serve as classification references and analogical explanations for the model's decisions.

\begin{figure*}[htbp]
\centering
\includegraphics[width=\textwidth]{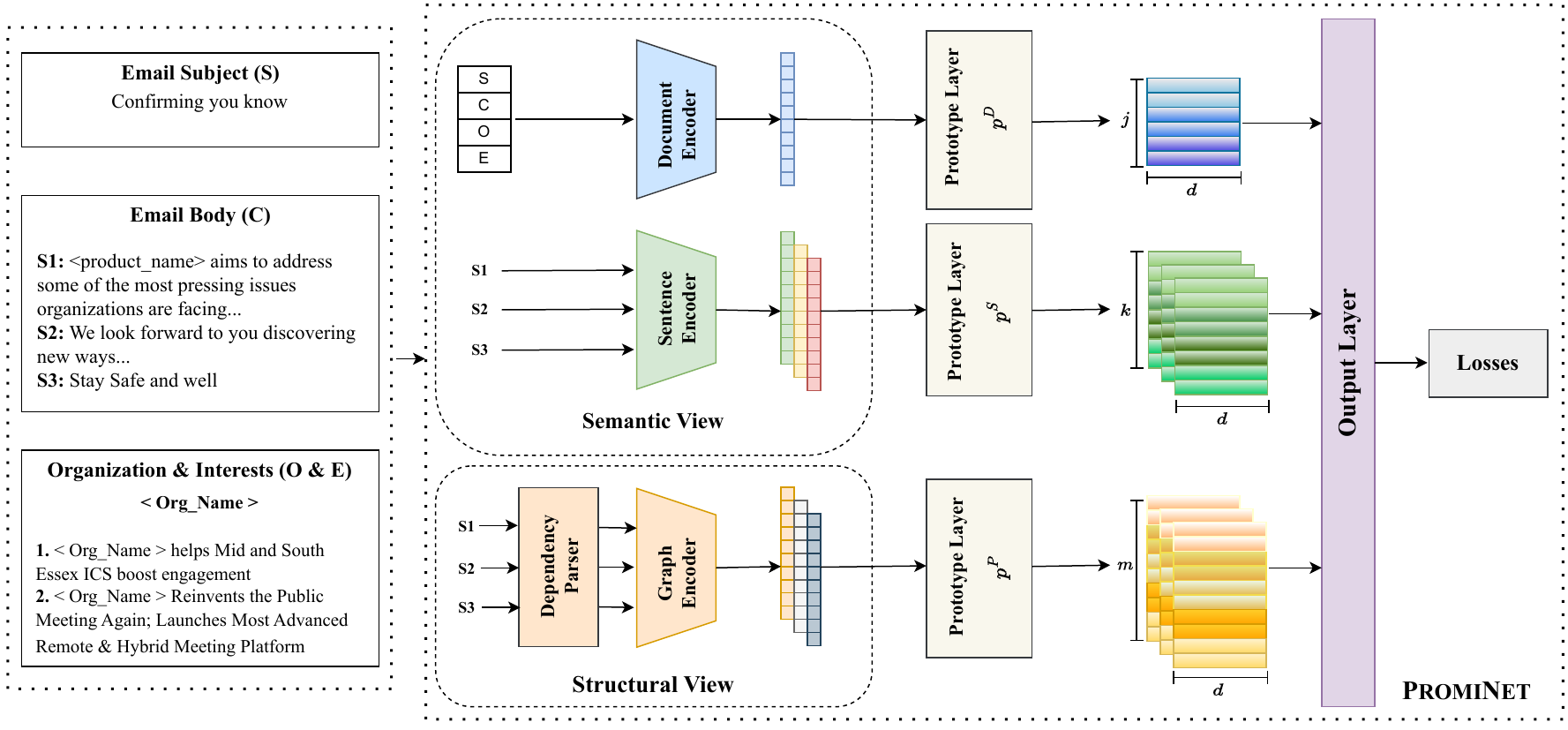}      
\caption{Illustration of our \textsc{ProMiNet} model. The model consists of an encoder and a prototype layer for each granularity $g$ (document ($D$), sentence ($S$) and phrase ($P$)) with two different views -- semantic \& structural.}
\label{fig:model_framework}
\end{figure*}

\section{Methodology}\label{methods}
In this section, we introduce \textsc{ProMiNet} model, that incorporates multi-view representations and prototype layers to develop a self-explainable email response prediction model. Our architectural choices prioritize two key factors: accuracy and interpretability. To ensure accurate email response predictions, our model leverages features derived from the email subject, body, and recipient information. It does so by employing a multi-view architecture that captures the interplay between different factors. The model extracts both structural and semantic information to comprehend the valuable cues pertaining to email persuasiveness and engagement. Moreover, our model is designed to be interpretable, offering insights into decision-making at various levels. Using the information from the multi-view representations, the model achieves interpretability through granular latent prototypes that serve as explanations for predictions. By considering both accuracy and interpretability, the model aims to strike a balance between making accurate predictions and providing transparent reasoning. In our \textsc{ProMiNet} model, we incorporate two main views, namely the Semantic view and the Structural view, to achieve our goal. We acquire embeddings at the document, sentence, and phrase-level by employing different components described in the subsequent subsections.

\subsection{Semantic View}
The Semantic view focuses on capturing features at both the document-level and sentence-level from email data. To extract document-level features, we employ a document encoder $(f^D)$ that considers the interaction between different elements such as the email subject (S), body/content (C), recipient organization (O), and their interests (E). These elements are separated by a special token ($[SEP]$), and we prepend the email with a token ($[CLS]$). By utilizing a pre-trained transformer-based encoder, the email data is transformed into token-level representations, where the $[CLS]$ token representation serves as the document-level embedding, $e^D$. For sentence-level features, a similar transformer-based sentence encoder $(f^S)$ is used to process each sentence within the email body. We add special tokens ($[CLS]$ and $[SEP]$) at the beginning and end of each sentence respectively. We denote the sentence-level embedding as $e^S$.

\subsection{Structural View}
The structural view emphasizes the importance of specific phrases in email engagement by examining the relationships between tokens or phrases within email sentences. By employing dependency parsing on the sentences, we create a graphical representation known as a dependency graph. The dependency graph comprises nodes representing tokens and links representing dependency relationships. These relationships are expressed as triples: ($v_{dep}$, $<rel>$, $v_{gov}$), where $v_{dep}$ and $v_{gov}$ denote the dependent and governing tokens, respectively; $<rel>$ refers to the dependency relationship between the tokens. To obtain phrase-level embeddings $e^P$, we extract dependency subgraphs from the sentences, focusing on dependencies like nominal subject (nsubj) and direct object (dobj) relative to the `ROOT' token. Utilizing a graph encoder $(f^P)$, we generate embeddings for each dependency subgraph, effectively capturing the structural information they convey.

\subsection{Prototype Layers}


In our approach, we utilize a prototype layer $p$ consisting of three sets of prototypes: $p^D \in \mathcal{R}^{j\times d}$ for latent document prototypes, $p^S \in \mathcal{R}^{k\times d}$ for sentence prototypes, and $p^P \in \mathcal{R}^{m\times d}$ for phrase prototypes, where $d$ is the dimension of the prototype embeddings (set identical to the dimensions of the output representations from the encoders) and $j, k, m$ refers to the number of prototypes associated with each granularity level. To guarantee effective representation of each class through learned prototypes at varying levels of granularity, we assign a fixed number of prototypes to each class. These prototypes are learned during the training process and represent groups of data instances, such as documents, sentences, or phrases, found in the training set. For each granularity level $g$, which can be either document (D), sentence (S), or phrase (P), the layer calculates the similarity between the granularity-specific embedding ($e^g$) and each trainable prototype. Formally,
\begin{align}
sim(p^{g}_i, e^{g}) = \log\left(\frac{||p^{(g)}_i-e^{g}||_2^2 + 1}{||p^{g}_i-e^{g}||_2^2 + \epsilon}\right)
\end{align}

Here, $p^g_i$ represents the $i^{th}$ prototype for granularity $g$, and it has the same dimension as the embedding ($e^g$). The similarity score decreases monotonically as the Euclidean distance $||p_i^g-e^g||_2$ increases, and it is always positive. For numerical stability, we set $\epsilon$ to a small value, specifically $1e-4$. We denote the computed similarity for each granularity level $g$ as $\mathcal{S}^g$.  

\subsection{Output Layer}
Finally, our model's output layer, denoted as $c$, includes a fully connected layer followed by a softmax layer to predict the likelihood of an email receiving a response. The prediction is determined by the weighted sum $\mathcal{S}^D + \lambda_1 \mathcal{S}^S + \lambda_2 \mathcal{S}^P$, which involves averaging the scores at the sentence and phrase levels with their weights denoted by $\lambda_1$ and $\lambda_2$, respectively.

\subsection{Learning Objectives}
We introduce different loss functions that ensure accuracy and interpretability.
For accuracy, we have cross entropy loss: 
\begin{equation}
L_{ce} = \frac{1}{n}\sum_{i=1}^n CE(c \circ p \circ f(x_i), y_i)
\end{equation}
where the output layer $c$ combines the information captured by different encoders ($f$) and prototype layers ($p)$ from multiple views at different granularity levels.
Drawing ideas from previous studies \cite{zhang2022protgnn,ming2019interpretable}, we introduce additional losses for prototype learning including: (a) diversity loss ($L_{div}$) that penalizes prototypes that are too similar to each other, (b) clustering loss ($L_{cls}$) that  ensures that each embedding (text or graph) is close to at least one prototype of its own class and (c) separation loss ($L_{sep}$) encourages each embeddings to be distant from prototypes not of its class. Formally,
\begin{align}
    L_{div} = \sum_{k=1}^C \sum_{\substack{q \neq r \\ p^g_q, p^g_r \in p}} \max(0, \cos(p_q^g, p_r^g) - \theta)\\
    L_{cls} = \frac{1}{n} \sum_{i=1}^n \min_{\substack{q: p^g_q \in p^g_{y_i}}} ||f^g(x_i) - p_q^g||_2^2\\
    L_{sep} = -\frac{1}{n} \sum_{i=1}^n \min_{\substack{q: p_q \notin p^g_{y_i}}} ||f^g(x_i) - p_q^g||_2^2
\end{align}
where $n$ is the total number of samples, $C$ is the number of classes, $\theta$ is the threshold of cosine similarity, and $\cos(\cdot, \cdot)$ measures the cosine similarity,  $p^g_{y_i}$ represents the set of prototypes belonging to class $y_i$ for granularity $g$. Finally, we use $L_1$ regularization as the sparsity loss ($L_{spa}$) to the output layer weights. The overall objective is:
\begin{equation}
    \mathcal{L}: = L_{ce} + \alpha L_{div} + \beta L_{cls} + \gamma L_{sep} + \delta L_{spa}
\end{equation}
where $\alpha, \beta, \gamma, \delta$ are the loss coefficients.

\subsection{Prototype Projection}
For improved interpretability, we project the latent prototypes onto the closest emails, sentences, or phrases from the training data. Each prototype's abstract representation is substituted with the nearest latent email, sentence, or phrase embedding in the training set that corresponds to its respective class of interest, measured by Euclidean distance. This conceptual alignment of prototypes with samples from the training set offers an intuitive and human-understandable interpretation of the prototypes associated with each class.

\section{Experimental Setup}
\subsection{Datasets}
\label{datasets}
Our framework is evaluated on two email datasets: the Enron corpus\footnote{\url{https://www.cs.cmu.edu/~enron/}} and the email marketing corpus. The Enron dataset, collected by the CALO Project, consists of $\sim500k$ emails from around 150 Enron Corporation employees. The email marketing corpus contains $\sim400k$ email data, including response details such as clicks, views, and replies from vendors. These emails were part of an email marketing program and only a subset of these emails get responded to. In order to handle the data imbalance,  we perform a random sampling to create a balanced split and conduct experiments over 5 runs. The dataset statistics and our other experimental settings for both the datasets are included in Appendix \ref{app:datasets}, \ref{app:hyperparam}.

\subsection{Baselines}
To demonstrate the effectiveness of our method, we compare our proposed $\textsc{ProMiNet}$ with transformer-based pretrained masked language models such as BERT-base \cite{kenton2019bert}, DistilBERT \cite{sanh2019distilbert}, RoBERTa \cite{liu2019roberta}, autoregressive language model like XLNet \cite{yang2019xlnet}, graph neural network-based TextGCN \cite{yao2019graph} that operates over a word-document heterogeneous graph, and prototype learning-based (ProSeNet \cite{ming2019interpretable} and ProtoCNN \cite{plucinski2021prototypical}) methods that learns to construct prototypes forsentences or phrases.







\begin{table}[htbp]
 \small
  \centering
  \small
  \begin{tabular}{c| c | c} 
\toprule
 \textbf{Methods} & \textbf{Enron} & \textbf{Email Marketing} \\
\midrule
 BERT-base & 83.9$_{\pm 2.9}$ & 78.9$_{\pm2.5}$ \\
 DistilBERT & 79.3$_{\pm2.6}$ & 73.6$_{\pm2.7}$ \\
 RoBERTa & 85.2$_{\pm3.0}$ & 79.5$_{\pm2.9}$ \\
 XLNet & \textbf{85.6$_{\pm3.4}$} & \textbf{80.2$_{\pm3.6}$} \\
 \midrule
 TextGCN & 80.9$_{\pm3.7}$ & 74.1$_{\pm3.4}$ \\
 \midrule
 ProSeNet & 82.1$_{\pm3.0}$ & 73.6$_{\pm3.2}$\\
 ProtoCNN & 83.6$_{\pm3.8}$ & 73.3$_{\pm3.6}$ \\
 \midrule
 \multicolumn{3}{c}{\textsc{ProMiNet Variants}} \\\hline
 BERT + GCN &  84.6$_{\pm3.3}$ & 81.1$_{\pm3.6}$ \\
 BERT + GAT &  85.2$_{\pm2.9}$ & 81.2$_{\pm2.6}$ \\
 RoBERTa + GCN & 87.8$_{\pm2.8}$ & \textbf{83.1$_{\pm3.2}$}$^\ast$ \\
 RoBERTa + GAT & 87.4$_{\pm3.4}$ & 82.6$_{\pm3.4}$ \\
 XLNet + GCN & 88.2$_{\pm3.2}$ & \textbf{83.1$_{\pm3.6}$}$^\ast$ \\
 XLNet + GAT & \textbf{88.6$_{\pm3.3}$}$^\ast$ & 82.6$_{\pm3.4}$ \\
\midrule
Improvement (\%) & 3.50 & 3.62 \\
\bottomrule
\end{tabular}
 \caption{Evaluation results on two email corpus. We report the weighted $F_1$ score (\%) \& SD based on 5 runs. Our method achieve statistically significant improvements over the closest baselines ($p < 0.01$).}
  \label{tab:classification_performance}
\end{table}

\subsection{Metrics}
We calculate both the macro $F_1$ and the weighted $F_1$-score to evaluate the performance of the proposed models in the context of email response prediction on both datasets. Nevertheless, we prioritize the weighted $F_1$-score as our primary evaluation metric due to the balanced class distributions in our data splits. Additionally, we present the mean and standard deviations of the $F_1$-score across five runs in Section \ref{sec:results}. Finally, we also perform a statistical analysis to assess the significance of the differences in $F_1$-scores between our proposed method and the nearest baselines using a paired t-test.
\section{Results \& Discussion}
\label{sec:results}
\subsection{Overall Performance Comparison}
Table~\ref{tab:classification_performance} summarizes our evaluation results. $\textsc{ProMiNet}$ consistently achieves the best performance on both datasets. Specifically, using XLNet encoder for texts and GAT encoder for dependency graphs, our model improves the weighted average $F_1$ score by 3.50\% for the Enron corpus. Similarly, with RoBERTa/XLNet encoder for texts and GCN encoder for dependency graphs, $\textsc{ProMiNet}$ improves the weighted average $F_1$ score by 3.62\% for the Email Marketing corpus. Compared to the other transformer-based models, $\textsc{ProMiNet}$ demonstrates performance improvements, indicating that incorporating dependency graphs enhances word connections and contextual meaning, leading to better overall performance. $\textsc{ProMiNet}$ outperforms TextGCN significantly, suggesting that considering local information, such as word order, in addition to global vocabulary information is crucial for accurate classification. Moreover, $\textsc{ProMiNet}$ surpasses prototype learning methods (ProSeNet and ProtoCNN), highlighting the importance of learning prototypes that capture both semantic and structural aspects.

\subsection{Ablation Study}
In our ablation experiments\footnote{Please refer to the Appendix for additional analyses, including information on Datasets, Hyperparameters, Ablation studies, Visualization, and Limitations. \label{fn:supp}}, we scrutinize the influence of various model components in $\textsc{ProMiNet}$. The amalgamation of XLNet, GAT, and prototype learning demonstrates the highest performance, underscoring their complementary attributes. Prototype-based models exhibit comparable performance to their non-interpretable counterparts. Moreover, the fusion of XLNet with prototype learning surpasses the combination of GAT and prototype learning, highlighting the significance of semantic information in text comprehension. This experiment not only illustrates the superiority of multi-view representations derived from both semantic and structural perspectives over models relying on embeddings from a single view, but also showcases that when coupled with prototype learning, $\textsc{ProMiNet}$ achieves the highest performance. 
We present comprehensive ablation studies that assess the impact of factors such as the number of prototypes, sensitivity to weights $(\lambda_1, \lambda_2)$, the contribution of various email metadata, and detailed error analyses. For further information, please refer to Appendix \ref{app:ablation}.

\begin{table}[htbp]
\small
  \centering
  \begin{tabular}{c| c | c} 
\toprule
 \textbf{Methods} & \textbf{Enron} & \textbf{Email Marketing} \\
 \midrule
 XLNet & 85.6$_{\pm 3.4}$ & 80.2$_{\pm3.6}$\\
 GAT & 81.4 $_{\pm3.1}$ & 78.1$_{\pm2.9}$\\
 XLNet + GAT & \textbf{88.2$_{\pm2.8}$} & \textbf{81.8$_{\pm3.0}$} \\
 \midrule
 XLNet + Prototypes & 86.2$_{\pm2.9}$ & 80.8$_{\pm3.1}$\\
 GAT + Prototypes & 82.9$_{\pm3.2}$ & 78.3$_{\pm3.6}$\\
 \midrule
 \makecell{XLNet + GAT + \\ Prototypes ($\textsc{ProMiNet}$)} & \textbf{88.6$_{\pm3.3}^\ast$} & \textbf{82.6$_{\pm 3.4}^\ast$} \\
\bottomrule

\end{tabular}
 \caption{Investigation of the Impact of Various Components in $\textsc{ProMiNet}$ on Both Datasets. We analyze different model variants to assess the influence of semantic and structural views, as well as prototype layers. }
  \label{tab:ablation_bg}
\end{table}

\subsection{Explanations for Prediction}

\subsubsection{Case study}
\begin{figure*}[htbp]
\centering
\includegraphics[width=0.95\textwidth]{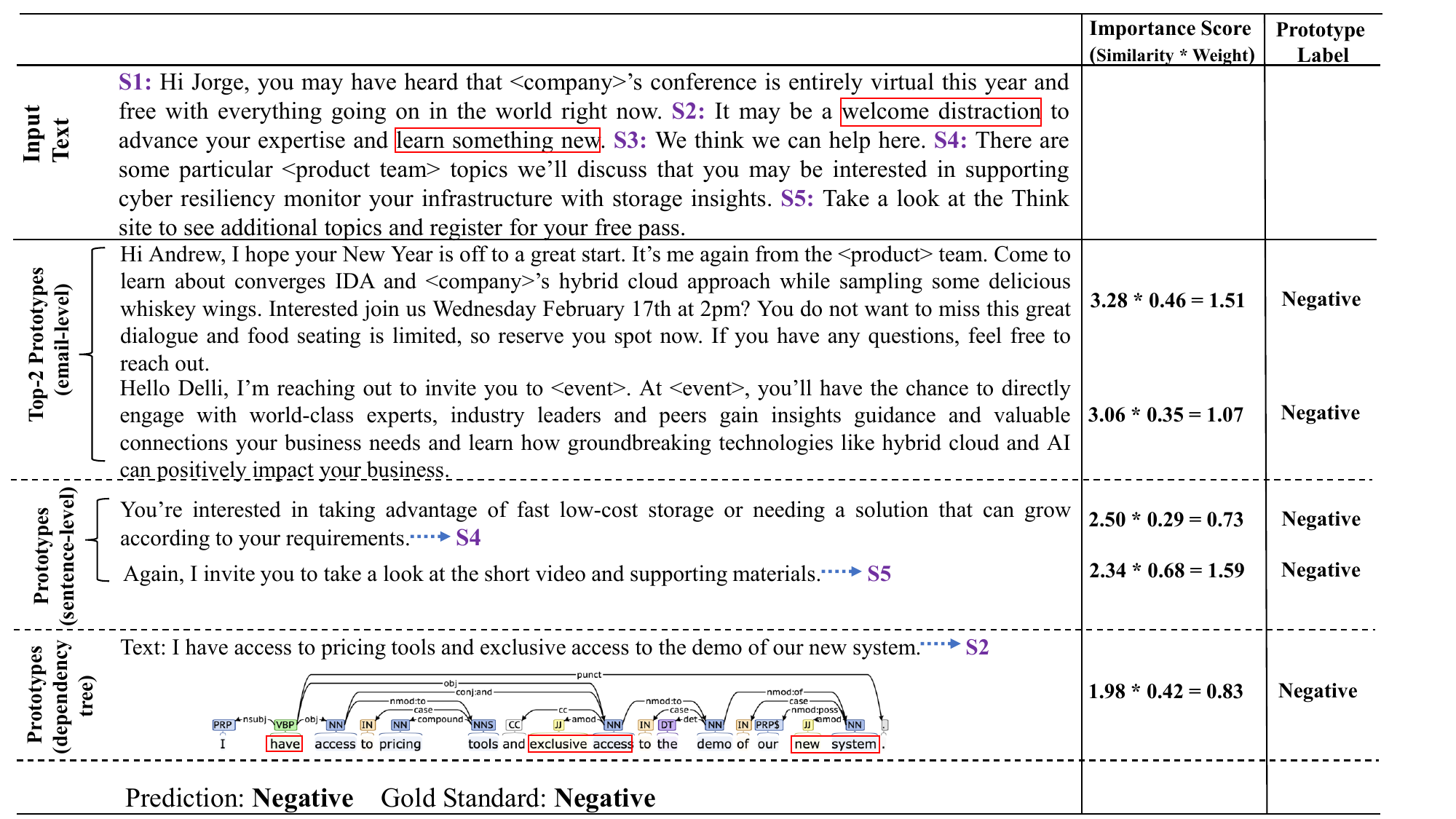}  
\caption{Example inputs and $\textsc{ProMiNet}$ prototypes for Email Marketing corpus. While classifying the input as negative (no response), the labels of prototypes are also negative. Due to space constraint, we only show a few prototypes with the largest weights.}
\label{fig:ibm_proto}
\end{figure*}


Figure~\ref{fig:ibm_proto} illustrates the reasoning process of $\textsc{ProMiNet}$ using an input example from the test set in the Email Marketing corpus. It showcases the most similar prototypes at the document, sentence, and phrase levels. The selected prototypes, along with their original labels, provide evidence for why the input example is classified as "negative". Two key observations emerge from the analysis: (a) All the learned prototypes associated with the input have the label "negative", consistent with the prediction of the input example; (b) The document-level prototypes exhibit similar topics to the input example, such as event invitations and basic event introductions. The sentence-level and phrase-level prototypes share similarities in terms of client interests, patterns, and grammatical relationships. We present similar analysis for an example from the Enron corpus in Appendix \ref{app:enron}.

\subsection{Suggest Edits based on Prototypes}

We utilize the attention mechanism from GAT to identify key phrases and important dependency relationships\footref{fn:supp} that contribute to the prediction. The words in a sentence are categorized into different types, such as nouns, verbs, adjectives, and adverbs. Since adjectives and nouns usually form key phrases, which are crucial, we focus on nouns and related words, considering their attention scores. Additionally, we use layer integrated gradients (LIG)~\cite{sundararajan2017axiomatic} and transformer-based embeddings to determine the importance of words in a sentence. After extracting the top-1 keyword/top-1 keyphrase for each sentence, we substitute keywords/keyphrases associated with prototypes with the label ``positive" (i.e., emails with response) for the keywords and key phrases of sentences in the test set with the label ``negative" (i.e., emails without response) to investigate whether there is a possibility to improve the ratio of ``positive" labels, that is, to improve the overall response rate. Here, the selected prototype emails share similar topics with the email to be edited. Otherwise, we randomly choose a prototype with response for edits. For an email, there are a few positions we consider editing: (1) email subjects; (2) email opening sentence/greeting (e.g., I hope you are doing well); (3) main contents of the email; (4) closing sentence (e.g., best regards). In our experience, we observe that using prototype-based edits of email subjects and main contents bring significant improvement of the overall email response rate on both datasets. For instance, the model captures the importance of creating a sense of urgency that improves the likelihood of receiving a response.  A sentence from an email labeled as ``negative" turns ``positive'' when the sentence  containing a phrase ``register for your free pass" is replaced with a prototype-based phrase ``get your free pass before the offer expires". Investigations on the impact of suggested edits on the effectiveness of our models are detailed in Appendix \ref{app:ablation}. 

\section{Conclusion}\label{discussion}
In this study, we introduced $\textsc{ProMiNet}$, a Prototype-based Multi-view Network that incorporates semantic and structural information from email data for interpretable email response prediction. $\textsc{ProMiNet}$ compared inputs to representative instances (prototypes) in the latent space to make predictions and offers prototypical explanations at the document, sentence, and phrase levels for enhanced human understanding. The evaluation on real-world email datasets demonstrates that $\textsc{ProMiNet}$  outperforms baseline models, achieving a significant improvement of approximately 3\% in $F_1$ score on both the Enron corpus and the Email Marketing corpus.  Our research contributes to enhancing sender-receiver communication and customer engagement in email interactions, filling a gap in understanding by considering email content and metadata. Future research directions involve addressing limitations such as time and historical interactions, handling unseen scenarios, improving interpretability, and balancing personalized content with prototypical information. These advancements will further propel the usage AI technqiues in  email marketing and communication.

\section*{Ethics Statement}
For this research, we utilized two distinct datasets. One of them comprises a publicly available collection, while the other involves IBM's internal email marketing corpus. It's important to note that we exclusively employed anonymized training data from the latter  \footnote{Although anonymized data was utilized for training and evaluation, in this paper, we have incorporated randomly generated names in email samples for the purpose of visualization and enhanced comprehension.}, ensuring the removal of any personally identifiable information. Furthermore, our methodology aims to enhance the interpretability of the email response prediction system, providing insights into the model's decision-making process at different levels of granularity without compromising proprietary or sensitive information. However, a noteworthy concern arises regarding the potential influence on user sentiments and actions in subtle ways, which could be interpreted as coercion. In such scenarios, the explanations provided through prototypes may inadvertently reveal biases or problematic training scenarios. This underscores the need for stringent guidelines and explainability, particularly in sensitive real-world contexts, to ensure that the model's predictions do not exert any harmful or ethically questionable influences on user decision-making. It's important to acknowledge that these risks are not unique to our methodology, but rather, they are pertinent to various AI techniques. This emphasizes the necessity for a consistent and vigilant review process and update of ethical standards and practices.

\bibliography{anthology,custom}
\bibliographystyle{acl_natbib}

\appendix
\newpage
\label{sec:appendix}
\setcounter{table}{0}
\renewcommand{\thetable}{A\arabic{table}}
\setcounter{figure}{0}
\renewcommand{\thefigure}{A\arabic{figure}}

\section{Dataset Details}
\label{app:datasets}
Summary statistics of the datasets are shown in Table~\ref{tab:data_stat}. Here, the striking difference in the ratio of ``response" and ``no response" samples between two email corpus is due to different email intents. The Enron corpus is pertaining to personal communication while the Email Marketing corpus is used for focused marketing campaigns.

\begin{table}[ht]
\small

  \centering
  \begin{tabular}{ccc}
    \toprule
    Datasets & Enron &  Email Marketing \\
    \midrule
    Total & 497,465 & 404,167 \\
    Response & 270,309 & 57,607  \\
    No Response &  227,156 &  346,560 \\
  \bottomrule
\end{tabular}
  \caption{Statistics of the datasets.}
  \label{tab:data_stat}
\end{table}

\subsection{Enron Corpus}
The Enron email dataset does not have explicit ``response" and ``no response" classes. Since ``reply" and ``forward" email threads appear in the original corpus, we categorize a single email as ``response"  as long as the original email contains ``reply" or ``forward" tag and extract only the portion of the email after the ``reply" or ``forward" tag. Otherwise, we categorize the single email as ``no response" and keep the entire email text. 

\subsection{Email Marketing Corpus}
We obtained this in-house corpus for research purposes. This dataset contains response information from clients in the form of clicks, views or replies. We label a single email as ``response"  as long as the original email is clicked, viewed, or replied at least once. Otherwise, we label the email as ``no response".

\section{Experimental Settings}
At encoder layer $f$, we use three variants of BERTs for text embedding, i.e., BERT-base, RoBERTa, and XLNet. Meanwhile, we use two variants of GNNs for subgraph embedding, i.e., GCN and GAT. Since the dataset is skewed, we perform a random downsample to create a balanced split and conduct experiments over 5 runs. We adopt the AdamW optimizer with a weight decay of 0.1. The hyperparameter search space for both datasets is included in Table \ref{tab:hp_space}. We perform random search for hyperparameter optimization. All of the experiments are conducted on four NVIDIA Tesla P100 GPUs.

\section{Hyperparameter Search Space}
\label{app:hyperparam}
\begin{table}[ht]
\small
  
  \centering
  \begin{tabular}{cc}

    \toprule
    Hyperparameters & Search Space\\
    \midrule
     Batch size & [16, 32, 64, 128] \\
     Learning rate & [$1\mathrm{e}{-5}, 2\mathrm{e}{-5}, 5\mathrm{e}{-5}$] \\
     Class weight for $l_{ce}$ & [0.2, 0.3, 0.4, 0.5] \\
     $j,k,m$ & [6, 10, 20, 30, 40, 50] \\
     $\theta$ & [0.2, 0.3, 0.4] \\
     $\alpha$ & [0.001, 0.005, 0.01, 0.015, 0.02] \\
     $\beta$ & [0.005, 0.01, 0.02, 0.05, 0.1] \\
     $\gamma$ & [0.001, 0.005, 0.01, 0.015, 0.02] \\
     $\delta$ & [0.001, 0.005, 0.01, 0.015, 0.02] \\
     $\lambda_1$ & [0.1, 0.3, 0.5, 0.7, 0.9] \\
     $\lambda_2$ & [0.1, 0.3, 0.5, 0.7, 0.9] \\
  \bottomrule
\end{tabular}
\caption{Hyperparameter search space of \textsc{ProMiNet} on both datasets. \label{tab:hp_space}}
\end{table}

\section{Ablation Studies}
\label{app:ablation}
\subsection{Effect of Email Components}
\begin{figure}[htbp]
\centering
\includegraphics[width=0.455\textwidth]{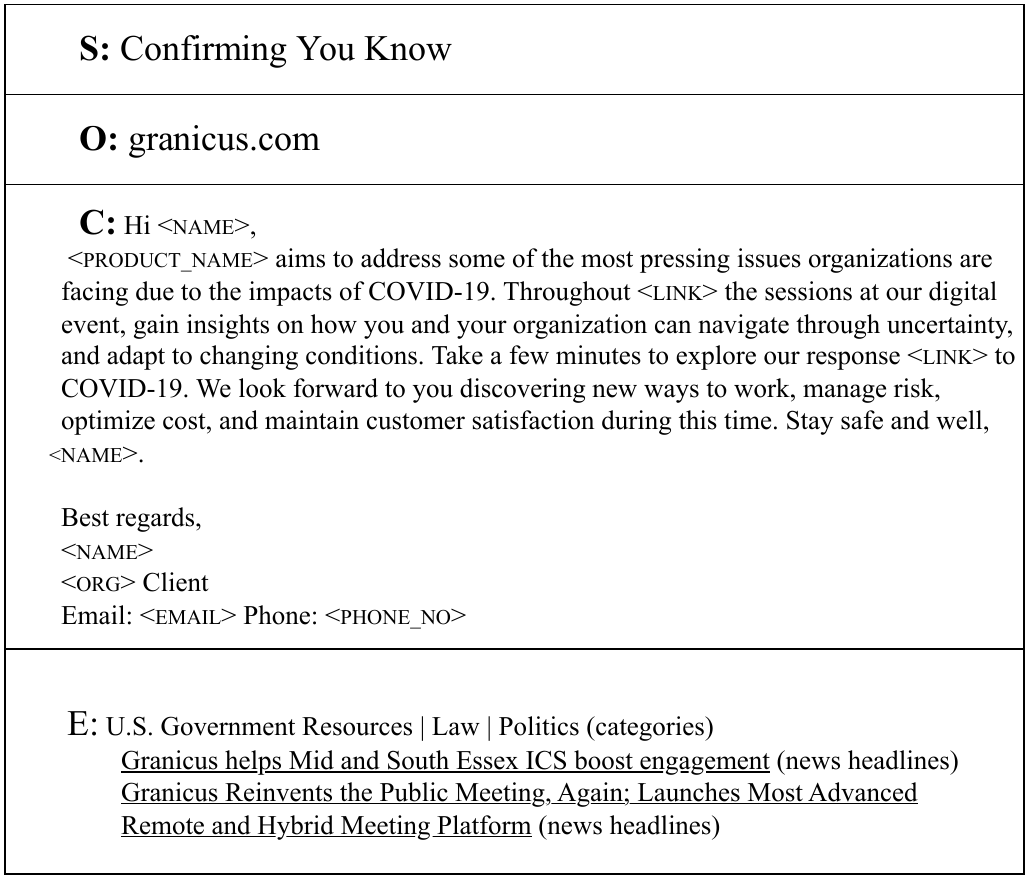}      
\caption{An example email from Email marketing corpus that contains subject (S), content (C), organization (O), and interests (E).}
\label{fig:ablation_email}
\end{figure}

We study the contribution of different email components as text inputs to the model's performance. In this study, we consider the subject, body text, and recipient's email organization as email composition components. Additionally, we utilize the AYLIEN news API\footnote{\url{https://aylien.com/}} to extract the interests of organizations. Our assumption is that the intent of an email may be associated with the recipient's organization's topic of interest. The API extracts news categories and headlines associated with the organization. For the Enron corpus, we evaluate the influence of the subject and body text only since all the recipients' email organizations in this corpus are from Enron. In the Email Marketing corpus, there are a considerable number of email recipient organizations for which the API is unable to extract interest information. In such cases, we leave the interests unknown. However, the goal of this experiment is to estimate the extent to which the interest information can boost our prediction performance. An example email from the Email Marketing corpus that contains all the pieces of information is provided in Figure \ref{fig:ablation_email}. This example helps in understanding the information contained in each part of the email before feeding it to the model. Based on the results presented in Table~\ref{tab:classification_performance}, we evaluate the contributions of email components using the $\textsc{ProMiNet}$ setting (XLNet + GAT) for the Enron corpus and the $\textsc{ProMiNet}$ setting (XLNet + GCN) for the Email Marketing corpus. We summarize the experimental results in Table~\ref{tab:ablation_email_comp} and make the following observations: The introduction of organization interests in the Email Marketing corpus shows marginal improvements in performance, confirming our assumption that there is an association between the intent of the sending email and the interests of the recipient's organization. The high standard deviation in performance when incorporating organization interests can be attributed to incomplete information. Despite these limitations, we observed some marginal improvement. A more in-depth analysis with complete information could yield significantly better results, but such investigation is beyond the scope of this paper and can be pursued in future research. When considering individual components of an email, the model's performance using body text as input outperforms the performance when using only the subject or email organization information. This finding highlights the significance of body texts in predicting email responses. The best performance is achieved when incorporating all three components—the subject, body text, and recipient's email organization. This indicates that each piece of information is valuable and contributes to performance gains. Overall, these observations emphasize the importance of considering multiple components and organization interests in improving the performance of email response prediction models.

\subsection{Error Analysis of Transformer-based/GNN Models}
Analyzing the error patterns of our Transformer-based/GNN models allows us to demonstrate the benefits provided by our $\textsc{ProMiNet}$ model. We focus on qualitatively examining email samples that are correctly classified by $\textsc{ProMiNet}$ but misclassified by other baseline models. For example, the sample email input provided in Figure~\ref{fig:ibm_proto} was misclassified by models that do not jointly model semantic and structural prototypes for response prediction. Additional analyses on models that solely utilize either semantic or structural prototypes are provided in Appendix~\ref{app:E}.

\subsubsection{Effects of Using only BERT/GNN}\label{app:E}
When using a combination of a transformer-based model and prototype learning, the input shown in Figure~\ref{fig:ibm_proto} is associated with the following top-2 email-level prototypes:
\begin{enumerate}
    \item  \textbf{Prototype 1}: ``Hi Phil, I hope this email finds you well. Just a quick line inviting you to attend <company\_name>'s online launch event storage made simple for all. During this event, you'll see how we are revolutionizing the entry enterprise storage space. If aah pharmaceuticals is challenged to deliver more with less budget it will be well worth your time attending.''

    \item \textbf{Prototype 2}: ``Hi John, hope this message finds you doing well today. My name is Nicholas Tompkins with <product\_name>, reaching out to personally invite you to an upcoming event. Did you know <company\_name> technology is simple innovative flexible fast and infused by AI. In this session, you will learn how we co-create solutions with you using flash systems virtualization, data protection cyber resiliency and business continuity strategies.''
\end{enumerate}
The most similar prototypes mapped to S4 and S5 are as follows:
\begin{itemize}
    \item Prototype mapped to S4: ``Hi Jack, do you have the need to refresh or add additional storage to your environment?''
    \item Prototype mapped to S5: ``Click here to register.''
\end{itemize}
The majority of prototypes associated with the input are labeled as "positive." However, the true label of the input is "negative." It is possible that although BERT captures the contextual information of an email, its ability to analyze dependencies and determine the grammatical structure of sentences is limited. Grammatical structures play a crucial role in enhancing sentence clarity and governing how words can be combined to form coherent sentences.
\begin{table}[htbp]
\centering
\resizebox{\columnwidth}{!}{
\begin{tabular}{c| c | c}
\hline
\makecell{Email \\ Components} & \makecell{Enron \\ $\textsc{ProMiNet}$ (XLNet + GAT)} & \makecell{Email Marketing \\ $\textsc{ProMiNet}$ (XLNet + GCN)}\\
 \hline
 \hline
 \textit{S} & 80.2 $\pm$ 3.6 & 76.9 $\pm$ 3.2 \\
 \textit{O} & ---- & 73.4 $\pm$ 2.9 \\
 \textit{C} & 85.1 $\pm$ 3.2 & 80.1 $\pm$ 3.4 \\
 \textit{S + O} & ---- & 78.4 $\pm$ 3.6 \\
 \textit{S + C} & \textbf{88.6 $\pm$ 3.3} & 82.6 $\pm$ 3.3 \\
 \textit{O + C} & ---- & 81.8 $\pm$ 3.7 \\
 \textit{S + O + C} & ---- & \textbf{83.1 $\pm$ 3.6} \\
 \textit{S + O + C + E} & ---- & \textbf{83.5 $\pm$ 4.1}$^\ast$ \\
\hline
\end{tabular}}
\caption{Effects of different email components as inputs of $\textsc{ProMiNet}$ on both datasets. The performance is evaluated via weighted average $F_1$ score (\%). Experiments are conducted with $5$ random initializations. The results are shown in the format of mean and standard deviation. Here, \textit{S}, \textit{O}, \textit{C}, and \textit{E} represent subject, organization, body text and organization interests, respectively.}
  \label{tab:ablation_email_comp}
\end{table}

\begin{figure*}[htbp]
\centering
\includegraphics[width=\textwidth]{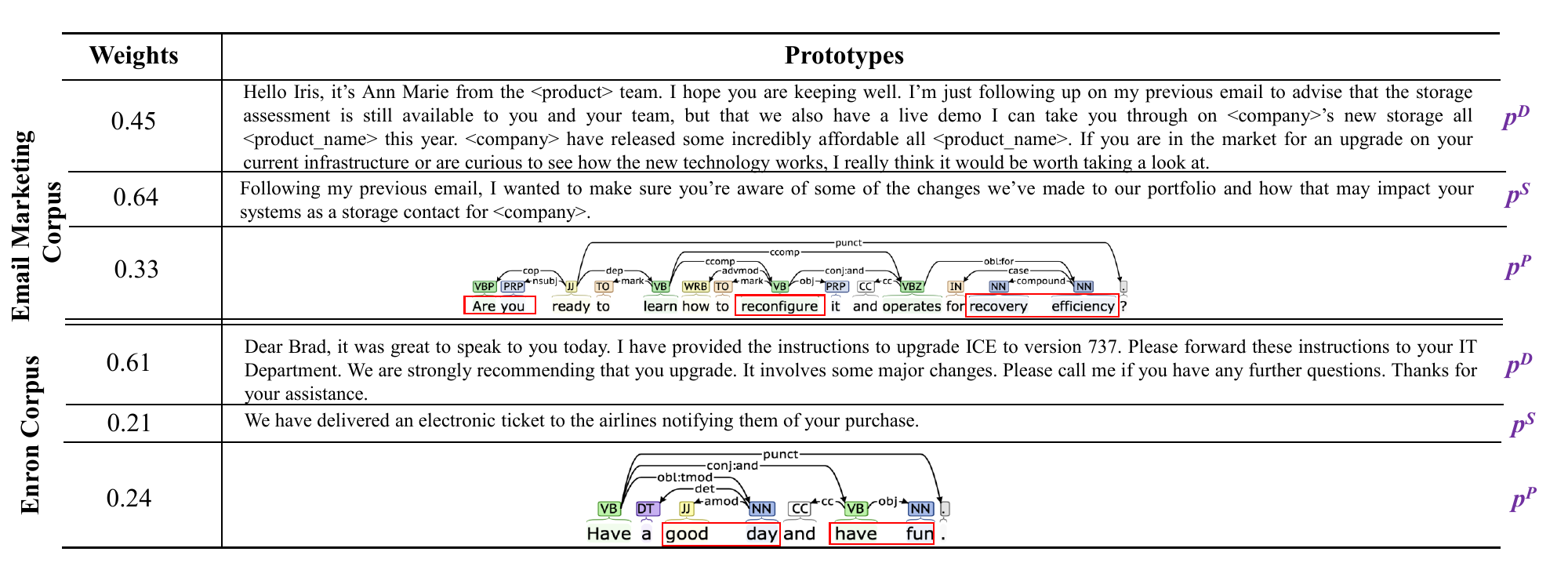}      
\caption{Visualization of three types of prototypes (i.e., document $(p^D)$, sentence $(p^S)$, phrase-level $(p^P)$) learned from the \textsc{ProMiNet} model on Enron Corpus and Email Marketing corpus.}
\label{fig:example_proto1}
\end{figure*} 

When employing a combination of GNN (Graph Neural Network) and prototype learning, we observe that the most similar prototype mapped to S2 can be seen in Figure~\ref{fig:ibm_proto}. However, there is a discrepancy between the label assigned to the prototype ("positive") and the label assigned to the sentence ("negative"). This mismatch suggests that the GNN component might lack the necessary information on text semantics to fully comprehend the content of the text.

This observation highlights the importance of investigating the interpretability capability of individual model components. In this case, it verifies the effectiveness of combining a transformer-based model, which excels at capturing contextual information, with a GNN, which is adept at capturing grammatical structures. The combination of these two components allows them to mutually influence and complement each other, resulting in a more comprehensive understanding of the input text.

\subsection{Effect of Suggested Edits}

We also investigate the impact of suggested edits on the effectiveness of our models. To simulate this, we conduct experiments where we make edits to the emails and observe the resulting changes in the ratio of ``positive'' labels. Table~\ref{tab:edit_improve} presents the ratios of original ``negative'' emails that are predicted as ``positive'' after making edits under different situations on both datasets. We employ a combination of XLNet and GAT for predictions and find that appropriate edits to email subjects and main content lead to improvements in the overall email response rate for both datasets. These improvements signify the potential of using prototypes to enhance the likelihood of generating favorable email responses.

\begin{table}[htbp]
\small  
  \centering
  \begin{tabular}{c| c | c} 
\hline
 Editing Positions  & Enron & Email Marketing \\
 \hline
 \hline
 Subjects & 1.4 $\pm$ 0.2 & 2.1 $\pm$ 0.3 \\
 Open sentence & 0.3 $\pm$ 0.0 & 0.9 $\pm$ 0.1\\
 Main contents & 1.9 $\pm$ 0.3 & 3.8 $\pm$ 0.5\\
 Closing sentence & 0.3 $\pm$ 0.0 & 0.4 $\pm$ 0.1 \\
\hline

\end{tabular}
\caption{Drop ratio of ``negative" labels after making edits on both datasets. }
  \label{tab:edit_improve}
\end{table}

\subsection{Effect of hyperparameters in $\textsc{ProMiNet}$}
We conducted a study to examine the impact of certain hyperparameters on model performance, specifically focusing on the number of prototypes and the addition weights.

\begin{figure}
\centering
\begin{subfigure}{.25\textwidth}
  \centering
  \includegraphics[width=\linewidth]{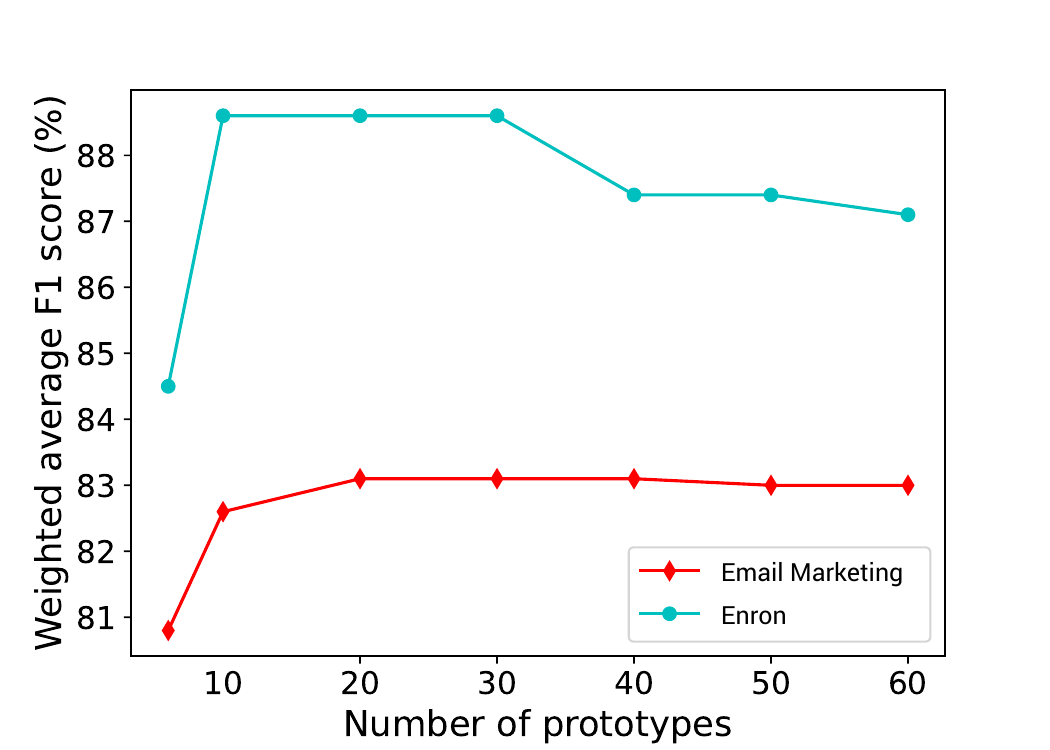}
  \caption{Effects of $j, k, m$.}
  \label{fig:num_proto}
\end{subfigure}%
\begin{subfigure}{.25\textwidth}
  \centering
  \includegraphics[width=\linewidth]{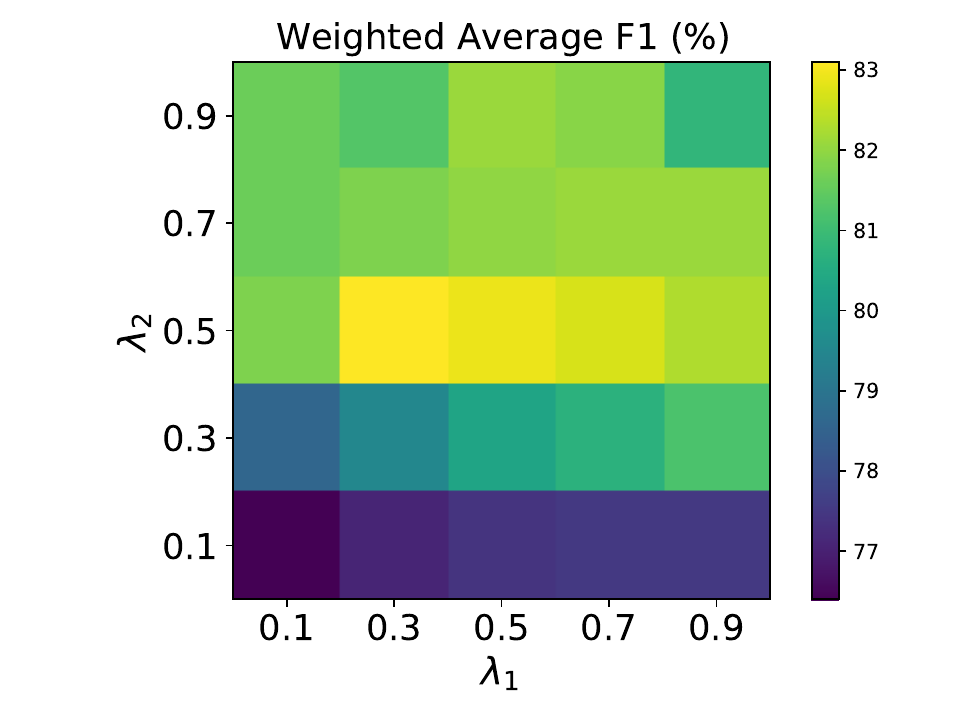}
  \caption{Effects of $\lambda_1, \lambda_2$.}
  \label{fig:relative_weight}
\end{subfigure}
\caption{Hyperparameter effects on performance.}
\label{fig:test}
\end{figure}

\textbf{Number of Prototypes (\textit{j,k,m}):} Figure~\ref{fig:num_proto} illustrates the relationship between the number of prototypes and the model performance, measured by the weighted average $F_1$ score, for both datasets. We observed that increasing the number of prototypes initially led to a significant improvement in performance. However, once the number of prototypes surpassed 20, the performance gains became less prominent, and in some cases, adding more prototypes even resulted in slightly worse performance. This phenomenon can be attributed to the increased complexity of the model, making it more challenging to train and comprehend. It demonstrates the trade-off between performance and interpretability. The optimal number of prototypes was found to be 20 for the Email Marketing corpus and 10 for the Enron corpus, as the model performance peaked at these values. 

\textbf{Addition Weight ($\bm{\lambda_1, \lambda_2}$):} The addition weights, $\lambda_1$ and $\lambda_2$, control the training balance among the three branches in our model. Figure~\ref{fig:relative_weight} presents the performance variations on the Email Marketing corpus when different combinations of $\lambda_1$ and $\lambda_2$ were used. The results demonstrate that the best performance was achieved when $\lambda_1$ was set to 0.3 and $\lambda_2$ was set to 0.5 in $\textsc{ProMiNet}$.

By investigating these hyperparameters, we gain insights into their effects on model performance, enabling us to optimize the performance and interpretability of our $\textsc{ProMiNet}$ model.
\begin{figure*}[htbp]
\centering
\includegraphics[width=\textwidth]{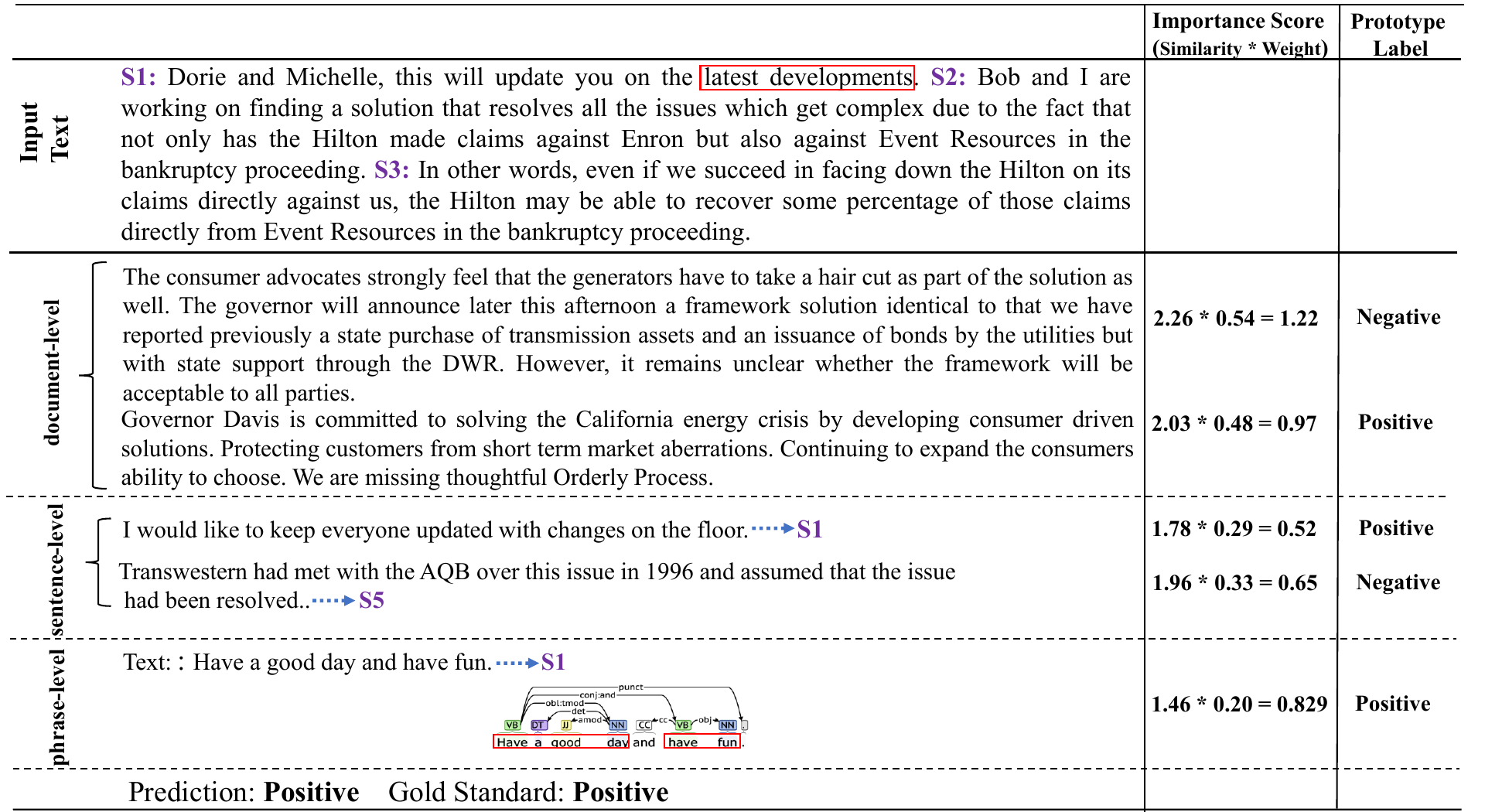}  
\caption{Example inputs and \textsc{ProMiNet} prototypes for Enron corpus. While classifying the input as positive (response), the majority of the prototype labels are also positive. Due to space constraint, we only show a few prototypes with the largest weights.
}
\label{fig:enron_proto}
\end{figure*}

\section{Case Study}\label{app:enron}

In Figure~\ref{fig:enron_proto}, we can examine the selected prototypes and their original labels, which serve as evidence for why the input example has been classified as positive. We can make two key observations:
\begin{itemize}
\item The majority of prototypes associated with the input have the label ``positive'', which aligns with the prediction of the input example being ``positive''.

\item The prototypes at the document-level share similar topics with the input example, specifically related to problem-solving. At the sentence-level, both S1 and its corresponding prototype discuss update notifications, while S2 and its prototype exhibit similar patterns. In terms of phrase-level prototypes, phrases extracted by S1 and its prototype share similar grammatical relationships, such as nominal subject (nsubj), adjectival modifier (amod), coordination (cc), and so on.
\end{itemize}

\section{Prototype Visualization}\label{app:D}
\begin{figure*}[t]
\centering
\includegraphics[width=0.9\textwidth]{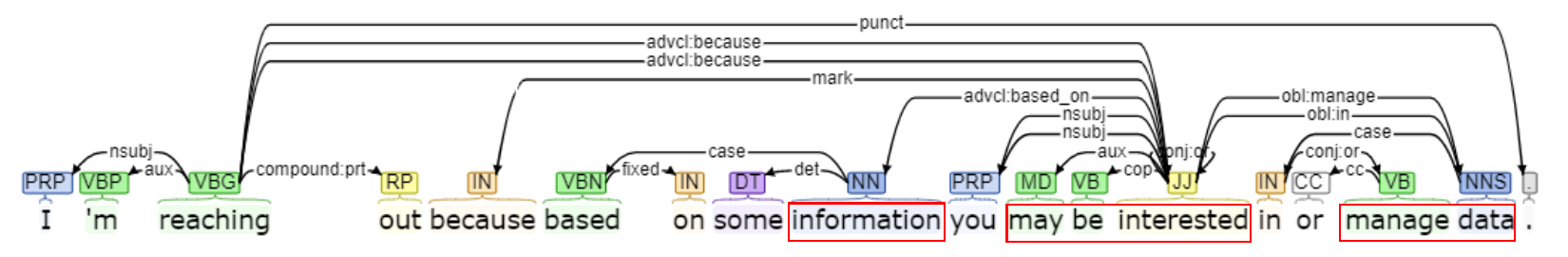}      
\caption{Most similar dependency subgraph prototype associated with S2 of input example in Figure~\ref{fig:ibm_proto} using only GNN.}
\label{fig:example_proto}
\end{figure*}

We provide prototype visualization, where each prototype is mapped to the latent representation of the most similar email in the training set. This mapping is facilitated by assigning static index numbers to each email or sentence from the same email during the model training phase. These index numbers enable us to visualize the prototypes later on. Figure~\ref{fig:example_proto} showcases some learned prototypes in a human-readable form for both datasets. The weight assigned to each prototype is derived from the fully connected layer. This diversity in different types of prototypes enhances our ability to provide explanations for prototype-based predictions.

\section{Limitations}
This work has several limitations that should be acknowledged. Firstly, the focus of this study is primarily on the text aspects of email data, disregarding factors such as time and historical interactions with customers. While this approach is suitable for the prediction task at hand, it overlooks potentially valuable contextual information that could impact email response behavior. Additionally, while prototypes are useful for the intended use case, there may be unseen scenarios or outliers that cannot be accurately mapped to examples in the training set, posing a challenge in dealing with such cases. Exploring alternative approaches to enhance interpretability and present explanations in a more user-friendly manner is an avenue for future research.  Furthermore, the prototype-based suggestion of edits presented in this work is a simulation experiment and may not capture the exact dynamics of real-time scenarios. The proposed shortcuts for improving model performance should be carefully considered to ensure alignment with actual email interactions. Lastly, using prototypical information in email composition runs the risk of generating templated emails with reduced personalization, even though personalization is known to be beneficial in email marketing \cite{sahni2018personalization}. Thus, addressing these limitations and exploring these areas of improvement could be the scope of future research.

\end{document}